\documentclass{article}
\usepackage{graphicx}
\usepackage{amssymb,amsmath}

\def\be{\begin{equation}}
\def\ee{\end{equation}}
\newcommand{\ff}[1]{{\mbox{\boldmath{$#1$}}}}
\def\x{\ff{x}}
\def\Dx{\Delta \x_i^t}
\def\Dv{\Delta \ff{v}_i^t}
\def\ra{\rightarrow}

\newcommand{\brak}[1]{\left\{  \begin{array}{lllllllll} #1 \end{array} \right. }

\def\citep#1{\cite{#1}}
\begin{document}

\title{Nature-Inspired Optimization Algorithms: Challenges and Open Problems}

\author{Xin-She Yang \\
School of Science and Technology, Middlesex University, London NW4 4BT, UK}

\date{}

\maketitle

\noindent {\bf Citation Detail}: \\
Xin-She Yang, Nature-Inspired Optimization Algorithms: Challenges and Open Problems, \\
\emph{Journal of Computational Science}, Article 101104, (2020). \\
https://doi.org/10.1016/j.jocs.2020.101104 \\ \hrule
\bigskip

%% Received 4 January 2020, Accepted 5 March 2020, Available online 6 March 2020. \\[10pt]

\abstract{
Many problems in science and engineering can be formulated as optimization problems, subject to complex nonlinear constraints. The solutions of highly nonlinear problems usually require sophisticated optimization algorithms, and traditional algorithms may struggle to deal
with such problems.
A current trend is to use nature-inspired algorithms due to their flexibility and effectiveness.
However, there are some key issues concerning nature-inspired computation and swarm intelligence. This paper provides an in-depth review of some recent nature-inspired algorithms with the emphasis on their search mechanisms and mathematical foundations. Some challenging issues are identified and five open problems are highlighted, concerning the analysis of algorithmic convergence and stability, parameter tuning, mathematical framework, role of benchmarking and scalability. These problems are discussed with the directions for future research.
}

%% The main text starts here
\section{Introduction}

Many real-world applications involve the optimization of certain objectives such as
the minimization of costs, energy consumption, environment and the maximization of performance, efficiency and sustainability. In many cases, the optimization problems that can be formulated are highly nonlinear with multimodal objective landscapes, subject to a set of complex, nonlinear constraints. Such problems are challenging to solve. Even with the ever-increasing power of modern computers, it is still impractical and not desirable to use simple brute force approaches. Thus, whenever possible, efficient algorithms are crucially important to such applications. However, efficient algorithms may not exist for most of the optimization problems in applications. Though there are a wide range of optimization algorithms such as gradient-based algorithms, the interior-point method and trust-region method, most of such algorithms are gradient-based and local search algorithms~\citep{Boyd2004,YangBook2014}, which means that the final solutions may depend on the initial starting points. In addition, the computation of derivatives can be computationally expensive, and some problems such as the objective with discontinuities may not have derivatives in certain regions.

A recent trend is to use evolutionary algorithms such as genetic algorithm (GA)~\citep{Holland1975},
and swarm intelligence (SI) based algorithms. In fact, a wide spectrum of SI-based algorithms have emerged in the last decades, including ant colony optimization (ACO)~\citep{Dorigo1992},
particle swarm optimization (PSO)~\citep{Kennedy1995PSO}, bat algorithm (BA)~\citep{YangBat2010},
firefly algorithm (FA)~\citep{Yang2009FA}, cuckoo search (CS)~\citep{YangDeb2009CS} and others \citep{YangBook2014}. These nature-inspired algorithms tend to be global optimizers, using a swarm of multiple, interacting agents to generate the search moves in the search space. Such global optimizers are typically simple, flexible and yet surprisingly efficient,
which have been shown in many applications and case studies (Kennedy et al., 2001; 
Engelbrecht, 2005; Yang et al., 2013; Yang, 2013; Yang, 2014). In the last three decades, significant progress has been and various applications have appeared. This paper will briefly summarize some of these important developments.

Despite the extensive studies and developments, there are still some important issues concerning swarm intelligence and nature-inspired algorithms. Firstly, there still lacks a unified mathematical framework to analyze these algorithms. Consequently, it lacks in-depth understanding how such algorithms may converge and how quickly they can converge. Secondly, there are many different algorithms and their comparison studies have mainly based on numerical experiments, and it is difficult to justify if such comparison is always fair. Thirdly, most of the applications in the literature concern small-scale problems, and it is not clear if such approaches can be directly applied to large-scale problems. Finally, it is not clear what are the conditions for the emergence of swarming and intelligence behaviour, even though the term `swarm intelligence' is used widely. All these mean that a systematical review and analysis is needed, and this paper is a preliminary attempt to analyze nature-inspired algorithms in a comprehensive and unified manner.

Therefore, this paper is organized as follows. Section 2 briefly review and summarize some of the recent SI-based algorithms, with the emphasis on their main characteristics. Section 3 focuses on the search mechanisms and their possible mathematical foundations. Section 4 attempts to highlight some of the main issues concerning nature-inspired algorithms from different perspectives, and outline some open problems and future research directions.

\section{Nature-Inspired Optimization Algorithms}

There are many nature-inspired algorithms in the current literature, it is estimated there are more than 100 different algorithms and variants (Reyes-Sierra and Coello Coello, 2006; Kennedy et al., 2001; Price et al., 2005; Fister et al., 2013; Yang and He, 2013; Alyasseri et al., 2018; Abdel-Basset and Shawky, 2019). It is not our intention to review all of them. Instead, our emphasis will be on the typical characteristics of algorithms and search mechanisms, and consequently we have selected only a few algorithms in our discussions here.  Though different algorithms can be described in different ways, it would be convenient for the discussions later if we can subdivide the descriptions of algorithms into two categories: procedure-based and equation-based.

\subsection{Procedure-Based Algorithms}

Though the genetic algorithm (GA) can have quite rigorous mathematical analyses~\cite{Holland1975,Goldberg1989}, it is mainly a procedure as an optimization algorithm.
Its main steps are carried out in an iterative manner, and its main procedure consists of three parts:
\begin{itemize}
\item Solution representations: A solution vector $\x$ to a $D$-dimensional problem is usually
represented or encoded as a binary string of a fixed length or a string of real numbers.

\item Solution modifications: Solutions can be modified by mutation or crossover. Mutation can be applied to a single solution at a single place or multiple places, while crossover is carried out over two parent solutions by mixing or swap relevant parts to form new solutions.

\item Solution selection: The fitness of a solution is evaluated, usually in terms of its objective value.  The selection of a solution among a population is carried out according to its fitness
    (higher values for maximization problems) and the best solutions are usually passed onto the next generation.

\end{itemize}
This iterative procedure is relative generic for many algorithms. For example, the evolutionary strategy (ES) can also fit into the above procedure, though crossover is not used in ES. In addition, the ant colony optimization (ACO)~\citep{Dorigo1992} can also fit into the above steps, though solution modifications are not by mutation or crossover. Chemical pheromone is used to represent the fitness of a solution, and the modification of solutions are by pheromone deposition and evaporation.

\subsection{Equation-Based Algorithms}
A vast majority of the recent nature-inspired algorithms for optimization are equation-based where all solution vectors $\x_i (i=1,2,...,n)$ are represented as a population set of $n$ solutions in a $D$-dimensional search space. In this sense, all different algorithms use the same type of vector representations of solutions.

In addition, the selection of the solutions is mainly based on their fitness values. The fittest solutions (higher objective values for maximization, or lower objective values for minimization) are most likely to be passed onto the next generation in the population. Though there some subtle form of selection, such as fitness-proportional elitism, the essence of solution selection is basically the same.

Consequently, the main differences among different nature-inspired optimization algorithms now are the ways of solution modifications, usually using different mathematical forms or search mechanisms. In general, a solution vector $\x_i^t$ at iteration or generation $t$ is a position vector, and the new solution $\x_i^{t+1}$ is generated by a modification increment or mutation vector $\Dx$. That is
\be \x_i^{t+1}=\x_i^t +\Dx, \ee
which dictates the main differences between different algorithms. Traditionally, this increment is a step size (or a step vector). In case of gradient-based algorithms such as the Newton-Raphson method, this step
is linked to the negative gradient
\be \Dx=-\eta \nabla f(\x), \ee
where $\nabla f$ is the gradient of the objective function $f(\x)$, and $\eta>0$ is the so-called learning parameter~\citep{Boyd2004,YangBook2014}.

In some nature-inspired algorithms, the modification in $\Dx$ is often related to the
increment of velocity modification $\Dv=\ff{v}_i^{t+1}-\ff{v}_i^t$ such as
\be \Dx  = \ff{v}_i^t \; \Delta t, \ee
where $\ff{v}_i^t$ and $\ff{v}_i^{t+1}$ are the velocity for the solution $i$ (particle, agent, etc.)
at iterations $t$ and $t+1$, respectively. Here, $\Delta t$ is the time increment. As all algorithms are iterative or time-discrete dynamical systems, the time step or increment $\Delta t$ is essentially the difference in the iteration counter $t$, which means that $\Delta t =1$
can be used for all these algorithms. Consequently, there is no need to worry about the units of these quantities and thus consider all quantities in the same units.

Now let us discuss the equations for modifying solutions in different algorithms.
\begin{itemize}

\item Differential evolution (DE): In differential evolution~\citep{Storn1997DE}, the main mutation is realized by
\be \x_i^{t+1}=\x_i^t + F(\x_j^t-\x_k^t), \ee
which can be written as
\be \x_i^{t+1}=\x_i^t + \Dx, \quad \Dx=F (\x_j^t-\x_k^t), \label{DE-equ-100} \ee
where $\x_i^t, \x_j^t$ and $\x_k^t$ are three distinct solution vectors from the population.
The parameter $F \in (0, 2)$ controls the mutation strength.

\item Particle swarm optimization (PSO)~\citep{Kennedy1995PSO}: The main inspiration of PSO comes from the swarming behaviour of birds and fish. The position and velocity of particle $i$ at any iteration or pseudo-time $t$ can be updated iteratively using
\be \ff{v}_i^{t+1}= \ff{v}_i^t  + \Dv, \label{pso-speed-100} \ee
\be \x_i^{t+1}=\x_i^t+ \Dx, \label{pso-pos-100} \ee
with
\be \Dx=\ff{v}_i^{t+1} \Delta t = \ff{v}_i^{t+1}, \ee
and \be
\Dv=\alpha \ff{\epsilon}_1 [\ff{g}^*-\x_i^t] + \beta \ff{\epsilon}_2 [\x_i^*-\x_i^t], \ee
where $\ff{\epsilon}_1$ and $\ff{\epsilon}_2$ are two uniformly distributed random numbers in [0,1]. Here,  $\ff{g}^*$ is the best solution of the population at iteration $t$, while $\x_i^*$ is the
individual best solution for particle $i$ among its search history up to iteration $t$.

\item Firefly algorithm (FA)~\citep{Yang2008Book,Yang2009FA}:  The main characteristics of FA are based on the attraction and flashing behaviour of tropical fireflies.
The position vector $\x_i$ of firefly $i$ at iteration $t$ is updated by
\be  \x_i^{t+1} =\x_i^t +\Dx, \label{FA-equ} \ee
and
\be \Dx= \beta_0 e^{-\gamma r^2_{ij} } (\x_j^t-\x_i^t) + \alpha \; \ff{\epsilon}_i^t, \ee
where $\beta_0>0$ is the attractiveness at zero distance, that is $r_{ij}=0$.
The scale-depending parameter $\gamma$  controls the visibility of fireflies,
while $\alpha$ essentially controls the strength of randomization in FA.

\item Bat algorithm (BA)~\citep{YangBat2010}: The main inspiration of BA is based on the
echolocation of microbats and the associated frequency-tuning characteristics in a range from $f_{\min}$ to $f_{\max}$, in combination with varying pulse emission rate and loudness~\citep{Altri1996,Colin2000Bat}.

The position of a bat is updated by
\be \ff{v}_i^{t} = \ff{v}_i^{t-1} +  \Dv, \label{f-equ-150} \ee
\be \x_i^{t}=\x_i^{t-1} + \Dx,  \label{f-equ-250} \ee
with
\be \Dv=(\x_i^{t-1} - \x_*) [f_{\min}+\beta (f_{\max}-f_{\min}) ],
\quad \Dx=\ff{v}_i^t \Delta=\ff{v}_i^t, \label{BA-equ-200} \ee
where $\x_*$ is the best solution among the population of $n$ bats, and $\beta$ is a random number in [0,1].

\item Cuckoo search (CS)~\citep{YangDeb2009CS}: CS was based on the aggressive reproduction
strategy of some cuckoo species and their interactions with host species such as warblers~\citep{Davies2011}. The eggs laid by cuckoos can be discovered and thus abandoned with a probability $p_a$, realized by a Heaviside step function $H$ with the use of a random number $\epsilon$ in [0,1]. The similarity of two eggs (solutions $\x_j$ and $\x_k$) can be roughly measured by their difference $(\x_j-\x_k)$. Thus, the position at iteration $t$
can be updated~\citep{YangDeb2014Rev} by
\be \x_i^{t+1}=\x_i^t +\Dx, \ee
where
\be \Dx= \alpha s \otimes H(p_a-\epsilon) \otimes (\x_j^t-\x_k^t). \label{CS-equ1} \ee
The step size $s$, scaled by a parameter $\alpha$ so as to limit its strength,
is drawn from a L\'evy distribution with an exponent $\lambda$~\citep{Pav2007Levy}.
The generation of this step size can be realized by some sophisticated algorithms such as the Mantegna's algorithm~\citep{Mantegna1994}.

\item Flower pollination algorithm (FPA) (Yang, 2012): The FPA was mainly based on the pollination processes and characteristics of flowering plants (Yang, 2012; Yang et al., 2013), including biotic and abiotic pollination as well as flower constancy~\citep{Waser1986}.
The solution vector $\x_i$ of a pollen particle $i$ can be simulated by
\be \x_i^{t+1}=\x_i^t + \Dx, \ee
and
\be \Dx=
\brak{\gamma L(\lambda) (\ff{g}_* - \x_i^t), & \textrm{ if } r<p, \\ \\
\epsilon (\x_j^t -\x_k^t), & \textrm{ otherwise.} }
 \ee
Here, $r$ is a uniformly distributed random number in [0, 1], and
$\gamma$ is a scaling parameter. $\ff{g}_*$ is the best solution found so far at iteration $t$. In the above equation, $L(\lambda)$ can be considered as a random number vector to be drawn from a L\'evy distribution with an exponent of $\lambda$~\citep{Pav2007Levy}.

\item Other Algorithms: There are many other nature-inspired algorithms, such as simulated annealing (Kirkpatrick et al., 1983), bacteria foraging optimization~\citep{Passino2002}, biogeography-based optimization~\citep{Simon2008BGO},
    gravitational search (Rashedi et al.,2009), charged particle system \citep{Kaveh2010ChSS},
    black-hole algorithm \citep{Hatamlou2012}, krill herd algorithm \\
    \citep{Gandomi2012Krill},
    eagle strategy \citep{Yang2012ES} and others. However, their main differences are in the ways of generating $\Dx$ and $\Dv$ from the population of the existing solutions.

    Most recently, new variants and applications are appearing regularly, including local ant system for allocating robot swarms~\citep{Khaluf2019}, hybrid ant and firefly algorithms~\citep{Goel2018}, usability feature selection by MBBAT~\citep{Gupta2017}, vehicle routing (Osaba et al., 2017; Osaba et al., 2019) and others (Alyasseri et al., 2018; Rango et al., 2018; Abdel-Basset and Shawky, 2019).

\end{itemize}

Though the expressions of $\Dx$ and $\Dv$ may be different and some algorithms do not use velocity at all, the detailed underlying search mechanisms may also be very different, even for seemingly similar expression of $\Dx$. For example, the mutation in Eq.~\eqref{DE-equ-100} of differential evolution seems to be similar to the form in Eq.~\eqref{pso-pos-100} for PSO. However, the former was done by random permutation, while the latter was done by a difference vector with perturbed directions using a uniform distribution. Similarly, the modification in Eq.~\eqref{BA-equ-200} in BA uses frequency tuning by varying frequencies, though it has some similarity to the mathematical term in Eq.~\eqref{pso-pos-100}. Therefore, in order to gain better insight, we should analyze the underlying search mechanisms and their mathematical or statistical foundations.

\subsection{Applications}

Before we discuss various search mechanisms in nature-inspired algorithms, let us briefly outline some of their recent applications. These algorithms have been applied in almost every area of science, engineering and industry, from engineering optimization (Cagnina et al, 2008; Deb et al., 2002;
Yang and Gandomi, 2012; Yang and Deb 2013; Gandomi and Yang, 2014; Chakri et al., 2017),
and deep learning (Papa et al., 2017) to the coordination of swarming robots (Rango et al., 2018; Palmieri et al., 2018) and the travelling salesman problem (Ouaarab et al., 2014; Osaba et al., 2016; Osaba et al., 2017; Osaba et al., 2019).

For comprehensive reviews, please refer to some recent review articles 
(Blum and Roli, 2003; Reyes-Sierra and Coello Coello, 2006; Senthilnath et al., 2011;
Yang and He, 2013; Yang and Deb, 2014; Alyasseri et al., 2018; Abdel-Basset and Shawky, 2019)
and books (Kennedy et al., 2001; Yang et al., 2013; Yang, 2013).

\section{Search Mechanisms and Theoretical Foundations}

Different algorithms usually use different search mechanisms, and these search moves are
often based on the underly probability distributions. Though solution modifications or perturbations
are largely part of mutation, we now focus on their statistical foundations and underlying mechanisms. Loosely speaking, we can put the ways of modifying or perturbing existing solutions into five categories: gradient-guided moves (GGM), random permutation (RP),
direction-based perturbations (DBP),  isotropic random walks (IRW),
and long-tailed, scale-free random walks (LTRW).

\begin{table}[ht]
\centering
\caption{Position and velocity modifications in algorithms. \label{PV-tab-100}}
\begin{tabular}{|l|l|l|}
\hline
Algorithm & Position increment $\Delta \x$ & Velocity increment $\Delta \ff{v}$ \\ \hline
Newton-Raphson & GGM & None \\
PSO & DBP & DBP \\
DE & RP, DBP & None \\
CS & RP, DBP, LTRW & None \\
SA & IRW & None \\
FA & DBP, IRW & None \\
BA & RP, DBP & RP, DBP \\
FPA & DBP, LTRW & None \\

\hline
\end{tabular}
\end{table}

Gradient-guided moves are mainly used in gradient-based optimization algorithms such as the Newton-Raphson method. The modification is parallel to the gradient direction, and the step length can be controlled by a learning parameter.

Random permutation tends to mix up a set of $n$ solutions, and then $k \ge 1$ solutions are randomly selected to generate new solutions. Random permutations are used in many algorithms, such as DE and FPA.

Direction-based perturbations are used in many algorithms with the term
of $(\ff{g}_*-\x_i)$ or $(\x_j-\x_k)$. The difference between any two vectors such as
$\x_j$ and $\x_k$ determines a direction, but this direction is then perturbed by multiplying by a uniformly distributed random number $\epsilon$. Thus, the actual directions of the moves are randomly distributed within a cone.

%%%%%%%%%%%%% a table for comparing search algorithms %%%%%%%%%%%%%%%%%%%%%

\begin{table}[ht]
\centering
\caption{Nature-inspired algorithms and their search characteristics.\label{Alg-tab-100}}
\begin{tabular}{|l|l|l|}
\hline
Algorithm & Probability  & Search \\
          & Distribution & Characteristics \\
\hline
PSO & Uniform & Guided search towards $\ff{g}_*$ \\
DE & Uniform  & Auto-scaling search \\
   &  Permutation & Random mutation  \\
CS & L\'evy flights & Self-similar search moves \\
   & Long-tailed & Scale-free search  \\
FA & Gaussian, uniform & Nonlinear attraction \\
BA & Uniform & Frequency-tuning \\
   &         & Fitness-dependent switching \\
FPA & Uniform & Scale-free search \\
    & L\'evy flights  & Jumps biased towards $\ff{g}_*$ \\
\hline
\end{tabular}
\end{table}

Random walks are a general framework for solution perturbations~\citep{YangHe2019}.
If we consider the current solution $\x_i^t$ as the current state $S_t$ at
time $t$, the next state (or solution) can be achieved by a local move $w_{t+1}$
\be S_{t+1}=S_t+w_{t+1}. \ee
Here, we use the non-bold form to denote the state in the $D$-dimensional space,
and perturbations can be done in a dimension by dimension manner.
Here, $w_{t+1}$ can be an array of random numbers to be drawn from the Gaussian distribution
\be w_{t+1} \sim N(0, 1), \ee
which means that the random walk becomes a Brownian motion. Here, the notation `$\sim$' emphasizes that the random steps should be drawn from the probability distribution described by the right-hand side of the equation. As the iteration time is discrete, the pseudotime counter $t$ can
be replaced by the number ($N=t$) of steps, which means that the average distance $d_N$ covered by a Brownian random walk is \be d_N \; \propto \; \sqrt{N}. \ee
This square-root law is a typical feature for many diffusion phenomena~\citep{YangHe2019}.
If the steps are drawn from Gaussian distributions, the random walks are isotropic random walks.

However, some probability density distributions can have a long tail, or a heavy tail.
If the steps are drawn from a heavy-tailed or long-tailed distribution, the random walks can
 become non-isotropic, long-tailed random walks or even scale-free random networks.
A good example of heavy-tailed distributions is
the Cauchy distribution
\be p(x, \mu, \gamma)=\frac{1}{\pi \gamma} \Big[\frac{\gamma^2}{(x-\mu)^2+\gamma^2} \Big], \quad -\infty < x < \infty, \ee
with two parameters $\mu$ and $\gamma$. Both its mean and variance are infinite or
undefined~\citep{Grindstead1997,Bhat2002}.

Another important long-tailed distribution is the L\'evy distribution, which has been used
in nature-inspired computation. L\'evy flights are a very special random walk whose steps are drawn from the L\'evy distribution.

The rigorous definition of L\'evy probability distribution can be tricky, involving an integral~\citep{Gutow2001,Pav2007Levy}
\be p(x) =\frac{1}{\pi} \int_0^{\infty} \cos(k x) e^{-\alpha |k|^{\beta}} dk, \quad (0 < \beta \le 2), \ee
where $\alpha>0$. The case of $\beta=1$ is equivalent to a Cauchy distribution, while $\beta=2$ leads to a normal distribution. For the practical purpose, we can use the following approximations for large steps ($s$)
\be L(s) \ra \frac{\alpha \; \beta \; \Gamma(\beta) \sin (\pi \beta/2)}{ \pi |s|^{1+\beta}}, \quad s \gg 0,  \ee
where $\Gamma(\beta)$ is the standard gamma function. Comparing the variance of the Brownian random walks, the variance or the distance covered by L\'evy flights increases much faster~\citep{YangBook2014}. The mean distance covered by the L\'evy flights after $N$ steps is \be d_N \; \propto \; N^{(3-\beta)/2}. \ee
This power-law feature is typically for super-diffusion phenomena~\citep{Pav2007Levy}.

By analyzing nature-inspired algorithms in great detail, we can summarize the search mechanisms and
their underlying statistical characteristics in Tables~\ref{PV-tab-100} and \ref{Alg-tab-100}.
Most algorithms modify their solution population directly without using velocities as shown in Table~\ref{PV-tab-100}. However, even for the same types of probability distributions, their role and
effects on search characteristics can be subtly different, and we summarize their search characteristics loosely in Table~\ref{Alg-tab-100}.

\section{Challenges and Open Problems}

Despite the effectiveness of nature-inspired algorithms and their popularity, there are still
many challenging issues concerning such algorithms, especially from theoretical perspectives.
Though researchers know the basic mechanisms of how such algorithms can work in practice, it is not quite clear why they work and under exactly what conditions. In addition, all nature-inspired algorithms have algorithm-dependent parameters, and the values of these parameters can affect the performance of the algorithm under consideration. However, it is not clear what the best values or settings are and how to tune these parameters to achieve the best performance. Furthermore,
though there are some theoretical analyses of some nature-inspired algorithms (Clerc and Kennedy, 2003; Chen et al., 2018), it still lacks a unified mathematical framework to analyze all algorithms to get in-depth understanding of their stability, convergence, rates of convergence and robustness.

In the rest of this paper, we will highlight five open problems concerning nature-inspired algorithms: mathematical framework for stability and convergence, parameter tuning, role of benchmarking, performance measures for fair comparison, and large-scale scalability.

\subsection{Mathematical Framework}

As almost all algorithms for optimization are iterative, traditional numerical analysis tends to use fixed-point theorems to see if it is possible to show the conditions for such theorems are satisfied. Basically, an iterative algorithm means that a new solution $\x_{k+1}$ can be obtained from
the current solution $\x_k$ by an algorithm $A$ with a parameter $\alpha$ or a set of parameters.
That is
\be \x_{k+1} =A(\x_k, \alpha), \label{Alg-iter-100} \ee
If we omit the bold font and use the standard notations in numerical analysis, we can write the above
equation simply as
\be x_{k+1}=A(x_k), \ee
without stating $\alpha$ explicitly.
Using the properties of function composites, we have
\be x_{k+1}=A^{k+1}(x_0)=(A \circ A^k)(x_0), \ee
where $x_0$ is the initial starting point.

From the well-known Banach fixed-point theorem~\citep{Granas2003,Khamsi2001Kirk}, we know that a fixed point $x_*$ can exist if $A(x_*)=x_*$ under
the condition that certain distance metric $\rho(.,.)$
\be \rho\big(A(x_i), A(x_j)\big) \le \theta \; \rho(x_i, x_j), \quad 0 \le \theta < 1, \ee
for all $x_i$ and $x_j$. This requires that $\rho$ is a shrinking or contracting metric.
If a fixed point exists, it is possible to approach this point iteratively via
\be \lim_{k \ra \infty} x_{k+1}=\lim_{k \ra \infty} A^{k+1}(x_0)=x_*. \ee However, for most nature-inspired algorithms, this condition may not be true at all~\citep{YangHe2019}.

Alternatively, we can view the iterative system such as Eq.~\eqref{Alg-iter-100} as a dynamical
system, which allows us to analyze its behaviour in the framework of dynamical system theory.
For example, the analysis of PSO was first carried out using dynamical system theory~\citep{Clerc2002}.

In case when an algorithm is linear in terms of its position or  solution vectors. It is possible to write the updating equations as a set of linear equations as  a time-discrete linear dynamical system
\be x_{k+1}=B x_k, \ee
where $B$ becomes a linear mathematical operator on $x_k$~\citep{Khalil1996}.
Its solution can be written as
\be x(k) =B^k x_0. \ee
The Lyapunov stability requires that all the $n$ eigenvalues $\lambda_i$ of $B$
must satisfy
\be |\lambda_i| \le 1. \ee
If $|\lambda_i|<1$ (without equality), the algorithm or system
becomes globally asymptotically stable.

Using a dynamical system framework, Chen et al. (2018) studied an
extended bat algorithm system
\begin{eqnarray}
v_{k+1}& = & -\zeta x_k+\theta v_k+\zeta g, \\
x_{k+1}& = & x_k+\theta v_k+\zeta g-\zeta x_k,
\end{eqnarray}
where $v_k$ and $x_k$ are the velocity and position of a bat at iteration $k$. Here, $g$ is
the best solution found by the current population. $\zeta$ and $\theta$ are two parameters.

The above two equations can be rewritten compactly as
\begin{equation}
Y_{k+1} = C Y_k+M g,
\end{equation}
where
\begin{equation}
Y_k=\begin{bmatrix}
x_k \\ v_k
\end{bmatrix}, \qquad
C=\begin{bmatrix}
1-\zeta & \theta \\-\zeta & \theta \\
\end{bmatrix}, \qquad
M=\begin{bmatrix}
\zeta \\ \zeta
\end{bmatrix}.
\end{equation}
Their analysis obtained some stability conditions for the parameters so that
\begin{equation}\begin{cases}
-1 \le \theta \le +1, \\
\zeta \ge 0, \\
2\theta-\zeta+2 \ge 0.
\end{cases}
\end{equation}
They also used numerical experiments confirmed such stability (Chen et al., 2018).

Another common way for analyzing the probabilistic convergence is to use Markov chain theory.
There are some extensive studies in this area. For example, the convergence of the genetic algorithm has been analyzed in terms of Markov chains (Suzuki, 1995; Aytug et al., 1996; Greenhalgh and Marshal, 2000; Gutjahr, 2010), and
this framework has been applied to analyze the cuckoo search algorithm (He et al., 2018)
and the bat algorithm (Chen et al., 2018). However, these studies have focused on the probabilistic convergence, but there still lacks information about the rate of convergence.

From the Markov chain theory~\citep{Grindstead1997,Ghate2008}, we know that the largest eigenvalue of a proper Markov chain is $\lambda_1=1$, it is believed that the second largest eigenvalue $0<\lambda_2<1$ controls the error variations $||E||$ or the rate of convergence
\be ||E|| \le  C (1-\lambda_2)^k, \ee
where  $C>0$ is a positive constant, which depends on the exact forms of the chains.
In principle, the chain should converge as $k \ra \infty$, but it can be very challenging to figure out this eigenvalue $\lambda_2$. There is almost no literature on this topic in the context of nature-inspired algorithms. Thus, an open problem in this area is as follows:

{\bf Open Problem 1.} How to build a unified framework for analyzing all nature-inspired algorithms mathematically, so as to obtain in-depth information about their convergence, rate of convergence, stability, and robustness?

As we have seen from the above, it seems that this framework may require a multidisciplinary approach to combine different mathematical, stochastic and numerical methods so that we can study algorithms from different perspectives.

\subsection{Parameter Tuning}

All nature-inspired algorithms have algorithm-dependent parameters, though the number of parameters can vary greatly. For traditional algorithms such as quasi-Newton methods, the tuning of a single parameter can have rigorous mathematical foundations (Boyd and Vandenberghe, 2004; Bertsekas et al., 2003; Chabert, 1999; Zdenek, 2009). However, for nature-inspired algorithms, the tuning is mainly empirical or by parametric studies~\citep{Eiben2011}.

Loosely speaking, an algorithm with $m$ parameters $\ff{p}_m$=($p_1, p_2, ..., p_m$) can be written schematically as
\be \x_{k+1}=A(\x_k | p_1, p_2, ..., p_m, \epsilon_1, ..., \epsilon_s), \ee
where $\epsilon_1, ..., \epsilon_s$ are $s$ different random numbers, which can be drawn from different probability distributions. To a certain degree, all these random numbers are drawn iteratively, thus the tuning of an algorithm will mainly be about the $m$ parameters. Thus, we can compactly write the above as
\be \x_{k+1}=A(\x_k, \ff{p}_m). \ee
If we use an algorithm $B$ to tune this algorithm, how was algorithm $B$ tuned initially? If we used another algorithm $C$ to tune algorithm $B$, how did we tune $C$ in the first place? In principle, we should use a well-tuned algorithm (or an algorithm without any parameters) to tune a new algorithm.
Thus, a key issue is how to tune an unknown algorithm properly?

Systematical brute-force tuning can be very time-consuming if the number of parameters is large. In addition, there is no guarantee that a well-tuned algorithm works well for one type of problems can work well for a different type of problems. It may be the case that parameter settings of an algorithm can be algorithm-dependent and problem-dependent if we want to maximize the overall performance.
In addition, even if an algorithm is tuned, its parameters become fixed after tuning. However,
there is no reason that we cannot vary the parameter during iterations. In fact, some studies showed that the variations of a parameter can be advantageous, which leads to self-adaptive variants.
For example, self-adaptive differential evolution seems to work better than its original version (Brest et al., 2006).

One way of tuning algorithms is to consider parameter tuning as a bi-objective process
so as to form a  self-tuning framework (Yang et al., 2013), where the algorithm to be tuned can be used to tune itself. This can still be a very time-consuming approach. Now we have another open problem concerning parameter tuning and parameter control.

{\bf Open Problem 2.} How to best tune the parameters of a given algorithm so that it can achieve
its best performance for a given set of problems? How to vary or control these parameters so as to maximize the performance of an algorithm?

\subsection{Role of Benchmarking and No-Free-Lunch Theorem}

For any new algorithm, especially a new nature-inspired algorithm, an important study is
to use benchmark functions to test how the new algorithm may perform, in comparison with other algorithms. Such benchmarking allows researchers to gain better understanding of the algorithm in terms of its convergence behaviour, stability and advantages as well as disadvantages.
However, the key question is what benchmarks should be used.

In the current literature, the benchmarking practice seems to use a set of test functions with different properties (such as mode shapes, separability and optima locality), there are many such benchmark test functions~\citep{Jamil2013} and some test suites designed by different conferences or research groups. As these functions are typically smooth, defined on some regular domains, they can serve some purpose, but such benchmarking is not actually much use in practice.
There are many reasons, but we only highlight two here. One reason is that these functions are often well-designed and sufficiently smooth, while real-world problems are much more diverse and can be very different from these test functions.  Another reason is that these test functions are typically unconstrained or with simple constraints on regular domains, while the problems in
real-world applications can have many nonlinear complex constraints and the domains can be formed by
many isolated regions or islands. Consequently, algorithms work well for test functions cannot work well in applications.

For an algorithm to be validated properly, testing and validation should include test functions with
irregular domains, subject to various constraints. For example, if an algorithm has been tested for all the benchmarks in the literature, there is no guarantee that it can still be effective to solve other problems such as the following newly designed, seemingly simple, two-dimensional function $f(x,y)$
\be f(x,y)=\sum_{i=-N}^N \sum_{j=-N}^N (|i|+|j|) \exp\Big[-a (x-i)^2 - a(y-j)^2\Big], \ee
in the domain of
\be |x-i|+|y-j| \le b=\frac{1}{a}, \quad \forall i, j, \ee
where $i,j$ are integers, $N=100$ and $a=10$. This function has $4N^2$ local peaks, but it has four highest peaks at four corners; however, its domain is formed by many isolated regions, or $4N^2=40000$ regions.

In the current literature, there are many different optimization algorithms. A key question naturally arises:  Which is the best one to use? Is there a universal tool that can be used to solve all or at least a vast majority of optimization problems? The simple truth is that there are no such algorithms. This conclusion has been formalized by Wolpert and Macready in 1997 in their influential work on the no-free-lunch (NFL) theorem \citep{Wolpert1997}. The NFL theorem states that if an algorithm $A$ can outperform another algorithm $B$ for finding the optima of some objective functions, then there are some other functions on which $B$ will outperform $A$. In other words, both $A$ and $B$ can perform equally well over all these functions if their performance is averaged over all possible problems or functions.

But the conclusions from most studies in terms of benchmarking seem to indicate that some algorithms are better than others. In practice, we know that some algorithms are indeed better than others, and  the quicksort for sorting numbers is indeed better than a method based on simple pair-wise comparison. Now how do we resolve this seemingly contradiction? The key to resolve this issue lies in the keywords `all' and `average'. In practical problem-solving,  we are always concerned with a particular set of problems, not all problems. We are also concerned with the actual individual performance of solving a particular problem, not the averaged performance over all problems.
As result, benchmarking using a finite set of algorithms and a finite set of functions becomes a zero-sum ranking problem~\citep{Joyce2018,YangHe2019}. On the other hand, recent studies seemed to indicate that free lunches may exist~\citep{Auger2010FL,Corne2003FL}, especially for continuous optimization \citep{Auger2010FL},  multi-objective optimization \citep{Corne2003FL} or co-evolution \citep{Wolpert2005}. Therefore, we now have an open problem concerning benchmarking.

{\bf Open Problem 3.} What types of benchmarking are useful? Do free lunches exist, under what conditions?

\subsection{Performance Measures}

For the benchmarking comparison of different algorithms, the conclusions can be influenced by the performance metrics used. To make a comparison, researchers have to select appropriate performance measures. In the current literature, comparison studies are mainly concerned with the accuracy, computational efforts, stability, and success rates.

For a given set of problems and a few algorithms, the algorithms obtained the most accurate solutions in comparison with some known or analytical solutions are considered better. Obviously, this will depend on the accuracy level and the stopping criteria used. Obviously, if one algorithm runs longer than others, even an ineffective algorithm may be able to obtain sufficiently good results if allowed to run much longer. Thus, to be fair, all algorithms should use the same computational efforts, which is usually realized by fixing the number of function evaluations.

An alternative approach is to use a fixed accuracy and compare the number of function calls or evaluations as a measure of computational costs. Algorithms with the smaller numbers of function evaluations are considered better. Even for the same number $N$ of function calls, there are different ways of using this fixed budget. If one algorithm first runs half of $N$ (or any other values) evaluations and select solutions, and then feed them into the run of the second half of $N$ evaluations, the performance may be different from the execution of the same algorithm with a single run of $N$ evaluations. Such different ways of implementing the same algorithm may lead to mixed conclusions.

Due to the statistical nature of nature-inspired algorithms, results are not exactly repeatable, and thus multiple runs are needed so as to get meaning statistics. Thus, some researchers use the best objective value obtained at the final iteration, together with their means, standard deviations and other statistics. This may give a fuller picture about the algorithms. Though a smaller standard derivation may indicate that the algorithm is more robust, but this may be linked to the problem under consideration. In addition, the ways of initializing the population and the probability distributions used in the algorithm may also influence such results, though it is not clear how initialization may exactly affect the final results.

Another measure used for comparison is the success rate. For multiple runs ($N_r$), there may be $N_s$ times that an algorithm is able to find the optimal solution, which means that the success rate is
the ratio $N_s/N_r$. However, this depends the way of how the success is defined. For a function $f(x)$ with a known optimal solution $x_*$ and the minimization objective $f_{\min}(x_*)$, the success can be defined by either $|x-x_*| \le \delta$ or $|f(x)-f_{\min}| \le \delta$ for a given small neighborhood such as $\delta=10^{-5}$. This can be two very different criteria if the landscape is relative flat.

Some studies use one or more performance measures, but it is not clear if the above performance measures are truly fair measures for a fair comparison.

{\bf Open Problem 4.} What are the most suitable performance metrics for fairly comparing all algorithms? Is it possible to design a unified framework to compare all algorithms fairly and rigorously.

\subsection{Algorithm Scalability}

From the application point of view, the most important indicator of the effectiveness of an algorithm is how efficiently it can solve a wide range of problems. Apart from the constraints posed by the no-free-lunch theorem, the efficiency of a given algorithm for a given type of problems can be largely affected by the size of problem instances. A well-known example is the travelling salesman problem (TSP) where a visitor is required to visit each city exactly once so as to minimize the overall distanced travelling through $n$ cities. For a small number of cities (say, $n \le 5$),
it is an easy problem. For a moderate or large $n$, this problem becomes an NP-hard problem~\citep{Cook1983,Arara2009,Goldreich2008}.
In this case, an algorithm that works well for small-scale problem instances cannot be scaled up to solve large-scale problems in a practically acceptable time scale.

Despite the diverse range of applications concerning nature-inspired algorithms and evolutionary algorithms, the problem sizes tend to be small or moderate, typically under several hundred parameters. It is not clear if these algorithms can be scaled up, by parallel computing, high-performance computing or cloud computing approaches.

{\bf Open Problem 5.} How to best scale up the algorithms that work well for small-scale problems to solve truly large-scale, real-world problems efficiently?

There are other open problems concerning nature-inspired algorithms, including how to achieve the optimal balance of exploitation and exploration, how to deal with nonlinear constraints effectively, and how to use these algorithms for machine learning and deep learning.

Nature-inspired computation is an active area of research. It is hoped that the above five open problems we have just highlighted can inspire more research in this area in the near future.

%%% References


\begin{thebibliography}{86}
\expandafter\ifx\csname natexlab\endcsname\relax\def\natexlab#1{#1}\fi
\providecommand{\bibinfo}[2]{#2}
\ifx\xfnm\relax \def\xfnm[#1]{\unskip,\space#1}\fi
%Type = Book
\bibitem[{Boyd and Vandenberghe(2004)}]{Boyd2004}
\bibinfo{author}{S.~P. Boyd}, \bibinfo{author}{L.~Vandenberghe},
  \bibinfo{title}{Convex Optimization}, \bibinfo{publisher}{Cambridge
  University Press}, \bibinfo{address}{Cambridge UK}, \bibinfo{year}{2004}.
%Type = Book
\bibitem[{Yang(2014)}]{YangBook2014}
\bibinfo{author}{X.-S. Yang}, \bibinfo{title}{Nature-Inspired Optimization
  Algorithms}, \bibinfo{publisher}{Elsevier Insight},
  \bibinfo{address}{London}, \bibinfo{year}{2014}.
%Type = Book
\bibitem[{Holland(1975)}]{Holland1975}
\bibinfo{author}{J.~Holland}, \bibinfo{title}{Adaptation in Nature and
  Artificial Systems}, \bibinfo{publisher}{University of Michigan Press},
  \bibinfo{address}{Ann Arbor, MI, USA}, \bibinfo{year}{1975}.
%Type = Book
\bibitem[{Dorigo(1992)}]{Dorigo1992}
\bibinfo{author}{M.~Dorigo}, \bibinfo{title}{Optimization, Learning, and
  Natural Algorithms}, \bibinfo{publisher}{Ph.D. Thesis, Politecnico di
  Milano}, \bibinfo{address}{Milan, Italy}, \bibinfo{year}{1992}.
%Type = Inproceedings
\bibitem[{Kennedy and Eberhart(1995)}]{Kennedy1995PSO}
\bibinfo{author}{J.~Kennedy}, \bibinfo{author}{R.~Eberhart},
\newblock \bibinfo{title}{Particle swarm optimization},
\newblock in: \bibinfo{booktitle}{Proceedings of the IEEE International
  Conference on Neural Networks}, \bibinfo{publisher}{IEEE},
  \bibinfo{address}{Piscataway, NJ, USA}, \bibinfo{year}{1995}, pp.
  \bibinfo{pages}{1942--1948}.
%Type = Inproceedings
\bibitem[{Yang(2010)}]{YangBat2010}
\bibinfo{author}{X.-S. Yang},
\newblock \bibinfo{title}{A new metaheuristic bat-inspired algorithm},
\newblock in: \bibinfo{editor}{C.~Cruz}, \bibinfo{editor}{J.~R. Gonz\'alez},
  \bibinfo{editor}{D.~A. Pelta}, \bibinfo{editor}{G.~Terrazas} (Eds.),
  \bibinfo{booktitle}{Nature Inspired Cooperative Strategies for Optimization
  (NISCO 2010)}, volume \bibinfo{volume}{284} of
  \textit{\bibinfo{series}{Studies in Computational Intelligence}},
  \bibinfo{publisher}{Springer}, \bibinfo{address}{Berlin, Germany},
  \bibinfo{year}{2010}, pp. \bibinfo{pages}{65--74}.
%Type = Inproceedings
\bibitem[{Yang(2009)}]{Yang2009FA}
\bibinfo{author}{X.-S. Yang},
\newblock \bibinfo{title}{Firefly algorithms for multimodal optimization},
\newblock in: \bibinfo{editor}{O.~Watanabe}, \bibinfo{editor}{T.~Zeugmann}
  (Eds.), \bibinfo{booktitle}{Proceedings of Fifth Symposium on Stochastic
  Algorithms, Foundations and Applications}, volume \bibinfo{volume}{5792},
  \bibinfo{publisher}{Lecture Notes in Computer Science, Springer},
  \bibinfo{year}{2009}, pp. \bibinfo{pages}{169--178}.
%Type = Inproceedings
\bibitem[{Yang and Deb(2009)}]{YangDeb2009CS}
\bibinfo{author}{X.-S. Yang}, \bibinfo{author}{S.~Deb},
\newblock \bibinfo{title}{Cuckoo search via l\'evy flights},
\newblock in: \bibinfo{booktitle}{Proceedings of World Congress on Nature \&
  Biologically Inspired Computing (NaBIC 2009)}, \bibinfo{publisher}{IEEE
  Publications}, \bibinfo{address}{USA}, \bibinfo{year}{2009}, pp.
  \bibinfo{pages}{210--214}.
%Type = Book
\bibitem[{Kennedy et~al.(2001)Kennedy, Eberhart, and Shi}]{Kennedy2001Book}
\bibinfo{author}{J.~Kennedy}, \bibinfo{author}{R.~C. Eberhart},
  \bibinfo{author}{Y.~Shi}, \bibinfo{title}{Swarm Intelligence},
  \bibinfo{publisher}{Academic Press}, \bibinfo{address}{London, UK},
  \bibinfo{year}{2001}.
%Type = Book
\bibitem[{Engelbrecht(2005)}]{Engel2005}
\bibinfo{author}{A.~P. Engelbrecht}, \bibinfo{title}{Fundamentals of
  Computational Swarm Intelligence}, \bibinfo{publisher}{Wiley},
  \bibinfo{address}{Hoboken, NJ, USA}, \bibinfo{year}{2005}.
%Type = Book
\bibitem[{Yang et~al.(2013)Yang, Cui, B, Gandom, and
  Karamanoglu}]{Yang2013Swarm}
\bibinfo{author}{X.-S. Yang}, \bibinfo{author}{Z.~H. Cui},
  \bibinfo{author}{X.~R. B}, \bibinfo{author}{A.~H. Gandom},
  \bibinfo{author}{M.~Karamanoglu}, \bibinfo{title}{Swarm Intelligence and
  Bio-Inspired Computaion: Theory and Applications},
  \bibinfo{publisher}{Elsevier}, \bibinfo{address}{London, UK},
  \bibinfo{year}{2013}.
%Type = Book
\bibitem[{Yang(2013)}]{Yang2013CSFA}
\bibinfo{author}{X.-S. Yang}, \bibinfo{title}{Cuckoo Search and Firefly
  Algorithm: Theory and Applications}, volume \bibinfo{volume}{516} of
  \textit{\bibinfo{series}{Studies in Computational Intelligence}},
  \bibinfo{publisher}{Springer}, \bibinfo{address}{Heidelberg, Germany},
  \bibinfo{year}{2013}.
%Type = Article
\bibitem[{Reyes-Sierra and Coello~Coello(2006)}]{Reyes2006Rev}
\bibinfo{author}{M.~Reyes-Sierra}, \bibinfo{author}{A.~C. Coello~Coello},
\newblock \bibinfo{title}{Multi-objective particle swarm optimizers: a survey
  of the state-of-the-art},
\newblock \bibinfo{journal}{Int J Comput Intell Res} \bibinfo{volume}{2}
  (\bibinfo{year}{2006}) \bibinfo{pages}{287--308}.
%Type = Book
\bibitem[{Price et~al.(2005)Price, Storn, and Lampinen}]{Price2005DE}
\bibinfo{author}{K.~Price}, \bibinfo{author}{R.~Storn},
  \bibinfo{author}{J.~Lampinen}, \bibinfo{title}{Differential Evolution: A
  Practical Approach to Global Optimization}, \bibinfo{publisher}{Springer},
  \bibinfo{address}{Berlin, Germany}, \bibinfo{year}{2005}.
%Type = Article
\bibitem[{Fister et~al.(2013)Fister, Fister~Jr, Brest, and Yang}]{Fister2013FA}
\bibinfo{author}{I.~Fister}, \bibinfo{author}{I.~Fister~Jr},
  \bibinfo{author}{J.~Brest}, \bibinfo{author}{X.-S. Yang},
\newblock \bibinfo{title}{A comprehensive review of firefly algorithms},
\newblock \bibinfo{journal}{Swarm and Evolutionary Computation}
  \bibinfo{volume}{13} (\bibinfo{year}{2013}) \bibinfo{pages}{34--46}.
%Type = Article
\bibitem[{Yang and He(2013)}]{YangHe2013BA}
\bibinfo{author}{X.-S. Yang}, \bibinfo{author}{X.-S. He},
\newblock \bibinfo{title}{Bat algorithm: literature review and applications},
\newblock \bibinfo{journal}{International Journal of Bio-Inspired Computation}
  \bibinfo{volume}{5} (\bibinfo{year}{2013}) \bibinfo{pages}{141--149}.
%Type = Incollection
\bibitem[{Alyasseri et~al.(2018)Alyasseri, Khader, Al-Betar, Awadallah, and
  Yang}]{Aly2018FPA}
\bibinfo{author}{Z.~A.~A. Alyasseri}, \bibinfo{author}{A.~T. Khader},
  \bibinfo{author}{M.~A. Al-Betar}, \bibinfo{author}{M.~A. Awadallah},
  \bibinfo{author}{X.-S. Yang},
\newblock \bibinfo{title}{Variants of the flower pollination algorithm: a
  review},
\newblock in: \bibinfo{editor}{X.-S. Yang} (Ed.),
  \bibinfo{booktitle}{Nature-Inspired Algorithms and Applied Optimization},
  \bibinfo{publisher}{Springer}, \bibinfo{address}{Cham}, \bibinfo{year}{2018},
  pp. \bibinfo{pages}{91--118}.
%Type = Article
\bibitem[{Abdel-Basset and Shawky(2019)}]{Abdel2019}
\bibinfo{author}{M.~Abdel-Basset}, \bibinfo{author}{L.~A. Shawky},
\newblock \bibinfo{title}{Flower pollination algorithm: a comprehensive
  review},
\newblock \bibinfo{journal}{Artificial Intelligence Review}
  \bibinfo{volume}{52} (\bibinfo{year}{2019}) \bibinfo{pages}{2533--2557}.
%Type = Book
\bibitem[{Goldberg(1989)}]{Goldberg1989}
\bibinfo{author}{D.~E. Goldberg}, \bibinfo{title}{Genetic Algorithms in Search,
  Optimization and Machine Learning}, \bibinfo{publisher}{Addison-Wesley},
  \bibinfo{address}{Reading, MA, USA}, \bibinfo{year}{1989}.
%Type = Article
\bibitem[{Storn and Price(1997)}]{Storn1997DE}
\bibinfo{author}{R.~Storn}, \bibinfo{author}{K.~Price},
\newblock \bibinfo{title}{Differential evolution: a simple and efficient
  heuristic for global optimization},
\newblock \bibinfo{journal}{J Global Optimization} \bibinfo{volume}{11}
  (\bibinfo{year}{1997}) \bibinfo{pages}{341--359}.
%Type = Book
\bibitem[{Yang(2008)}]{Yang2008Book}
\bibinfo{author}{X.-S. Yang}, \bibinfo{title}{Nature-Inspired Metaheurisic
  Algorithms}, \bibinfo{publisher}{Luniver Press}, \bibinfo{address}{Bristol,
  UK}, \bibinfo{year}{2008}.
%Type = Book
\bibitem[{Altringham(1996)}]{Altri1996}
\bibinfo{author}{J.~D. Altringham}, \bibinfo{title}{Bats: Biology and
  Behaviour}, \bibinfo{publisher}{Oxford University Press},
  \bibinfo{address}{Oxford, UK}, \bibinfo{year}{1996}.
%Type = Book
\bibitem[{Colin(2000)}]{Colin2000Bat}
\bibinfo{author}{T.~Colin}, \bibinfo{title}{The Variety of Life},
  \bibinfo{publisher}{Oxford University Press}, \bibinfo{address}{Oxford, UK},
  \bibinfo{year}{2000}.
%Type = Article
\bibitem[{Davies(2011)}]{Davies2011}
\bibinfo{author}{N.~B. Davies},
\newblock \bibinfo{title}{Cuckoo adaptations: trickery and tuning},
\newblock \bibinfo{journal}{Journal of Zoology} \bibinfo{volume}{284}
  (\bibinfo{year}{2011}) \bibinfo{pages}{1--14}.
%Type = Article
\bibitem[{Yang and Deb(2014)}]{YangDeb2014Rev}
\bibinfo{author}{X.-S. Yang}, \bibinfo{author}{S.~Deb},
\newblock \bibinfo{title}{Cuckoo search: recent advances and applications},
\newblock \bibinfo{journal}{Neural Computing and Applications}
  \bibinfo{volume}{24} (\bibinfo{year}{2014}) \bibinfo{pages}{169--174}.
%Type = Article
\bibitem[{Pavlyukevich(2007)}]{Pav2007Levy}
\bibinfo{author}{I.~Pavlyukevich},
\newblock \bibinfo{title}{L\'evy flights, non-local search and simulated
  annealing},
\newblock \bibinfo{journal}{Journal of Computational Physics}
  \bibinfo{volume}{226} (\bibinfo{year}{2007}) \bibinfo{pages}{1830--44}.
%Type = Article
\bibitem[{Mantegna(1994)}]{Mantegna1994}
\bibinfo{author}{R.~N. Mantegna},
\newblock \bibinfo{title}{Fast, accurate algorithm for numerical simulation of
  l\'evy stable stochastic process},
\newblock \bibinfo{journal}{Physical Review E} \bibinfo{volume}{49}
  (\bibinfo{year}{1994}) \bibinfo{pages}{4677--83}.
%Type = Incollection
\bibitem[{Yang(2012)}]{Yang2012FPA}
\bibinfo{author}{X.-S. Yang},
\newblock \bibinfo{title}{Flower pollination algorithm for global
  optimization},
\newblock in: \bibinfo{editor}{J.~Durand-Lose}, \bibinfo{editor}{N.~Jonoska}
  (Eds.), \bibinfo{booktitle}{Unconventional Computation and Natural
  Computation (UCNC 2012}, volume \bibinfo{volume}{7445},
  \bibinfo{publisher}{Springer}, \bibinfo{address}{Berlin Heidelberg, Germany},
  \bibinfo{year}{2012}, pp. \bibinfo{pages}{240--249}.
%Type = Article
\bibitem[{Yang et~al.(2013)Yang, Karamanoglu, and He}]{Yang2013FPA}
\bibinfo{author}{X.-S. Yang}, \bibinfo{author}{M.~Karamanoglu},
  \bibinfo{author}{X.~S. He},
\newblock \bibinfo{title}{Multi-objective flower algorithm for optimization},
\newblock \bibinfo{journal}{Procedia Computer Science} \bibinfo{volume}{18}
  (\bibinfo{year}{2013}) \bibinfo{pages}{861--868}.
%Type = Article
\bibitem[{Waser(1986)}]{Waser1986}
\bibinfo{author}{N.~M. Waser},
\newblock \bibinfo{title}{Flower constancy: definition, cause and measurement},
\newblock \bibinfo{journal}{American Naturalist} \bibinfo{volume}{127}
  (\bibinfo{year}{1986}) \bibinfo{pages}{596--603}.
%Type = Article
\bibitem[{Kirkpatrik et~al.(1983)Kirkpatrik, GEllat, and Vecchi}]{Kirk1983SA}
\bibinfo{author}{S.~Kirkpatrik}, \bibinfo{author}{C.~D. GEllat},
  \bibinfo{author}{M.~P. Vecchi},
\newblock \bibinfo{title}{Optimization by simulated annealing},
\newblock \bibinfo{journal}{Science} \bibinfo{volume}{220}
  (\bibinfo{year}{1983}) \bibinfo{pages}{671--680}.
%Type = Article
\bibitem[{Passino(2002)}]{Passino2002}
\bibinfo{author}{K.~Passino},
\newblock \bibinfo{title}{Biomimicry of bacterial foraging for distributed
  optimization and control},
\newblock \bibinfo{journal}{IEEE Control Systems} \bibinfo{volume}{22}
  (\bibinfo{year}{2002}) \bibinfo{pages}{52--67}.
%Type = Article
\bibitem[{Simon(2008)}]{Simon2008BGO}
\bibinfo{author}{D.~Simon},
\newblock \bibinfo{title}{Biogeography-based optimization},
\newblock \bibinfo{journal}{IEEE Transactions on Evolutionary Computatio}
  \bibinfo{volume}{12} (\bibinfo{year}{2008}) \bibinfo{pages}{702--713}.
%Type = Article
\bibitem[{Rashedi et~al.(2009)Rashedi, Nezamabadi-Pour, and
  Saryazdi}]{Rashedi2009GSA}
\bibinfo{author}{E.~Rashedi}, \bibinfo{author}{H.~H. Nezamabadi-Pour},
  \bibinfo{author}{S.~Saryazdi},
\newblock \bibinfo{title}{Gsa: a gravitational search algorithm},
\newblock \bibinfo{journal}{Information sciences} \bibinfo{volume}{179}
  (\bibinfo{year}{2009}) \bibinfo{pages}{2232--2248}.
%Type = Article
\bibitem[{Kaveh and Talatahari(2010)}]{Kaveh2010ChSS}
\bibinfo{author}{A.~Kaveh}, \bibinfo{author}{S.~Talatahari},
\newblock \bibinfo{title}{A novel heuristic optimization method: charged system
  search},
\newblock \bibinfo{journal}{Acta Mechanica} \bibinfo{volume}{213}
  (\bibinfo{year}{2010}) \bibinfo{pages}{267--289}.
%Type = Article
\bibitem[{Hatamlou(2012)}]{Hatamlou2012}
\bibinfo{author}{A.~Hatamlou},
\newblock \bibinfo{title}{Black hole: A new heuristic optimization approach for
  data clustering},
\newblock \bibinfo{journal}{Information Sciences} \bibinfo{volume}{222}
  (\bibinfo{year}{2012}) \bibinfo{pages}{175--184}.
%Type = Article
\bibitem[{Gandomi and Alavi(2012)}]{Gandomi2012Krill}
\bibinfo{author}{A.~Gandomi}, \bibinfo{author}{A.~Alavi},
\newblock \bibinfo{title}{Krill herd: a new bio-inspired optimization
  algorithm},
\newblock \bibinfo{journal}{Communications in Nonlinear Science and Numerical
  Simulatio} \bibinfo{volume}{17} (\bibinfo{year}{2012})
  \bibinfo{pages}{4831--4845}.
%Type = Article
\bibitem[{Yang and Deb(2012)}]{Yang2012ES}
\bibinfo{author}{X.-S. Yang}, \bibinfo{author}{S.~Deb},
\newblock \bibinfo{title}{Two-stage eagle strategy with differential
  evolution},
\newblock \bibinfo{journal}{International Journal of Bio-Inspired Computation}
  \bibinfo{volume}{4} (\bibinfo{year}{2012}) \bibinfo{pages}{1--5}.
%Type = Article
\bibitem[{Khaluf et~al.(2019)Khaluf, Vanhee, and Simoens}]{Khaluf2019}
\bibinfo{author}{Y.~Khaluf}, \bibinfo{author}{S.~Vanhee},
  \bibinfo{author}{P.~Simoens},
\newblock \bibinfo{title}{Local ant system for allocating robot swarms to
  time-constrained tasks},
\newblock \bibinfo{journal}{Journal of Computaitonal Science}
  \bibinfo{volume}{31} (\bibinfo{year}{2019}) \bibinfo{pages}{33--44}.
%Type = Article
\bibitem[{Goel and Maini(2018)}]{Goel2018}
\bibinfo{author}{R.~Goel}, \bibinfo{author}{R.~Maini},
\newblock \bibinfo{title}{A hybrid of ant colony and firefly algoirthms (hafa)
  for solving vehicle routing problems},
\newblock \bibinfo{journal}{Journal of Computational Science}
  \bibinfo{volume}{25} (\bibinfo{year}{2018}) \bibinfo{pages}{28--37}.
%Type = Article
\bibitem[{Gupta and Ahlawat(2017)}]{Gupta2017}
\bibinfo{author}{D.~Gupta}, \bibinfo{author}{A.~Ahlawat},
\newblock \bibinfo{title}{Usability feature selection via mbbat: A novel
  approach},
\newblock \bibinfo{journal}{Journal of Computational Science}
  \bibinfo{volume}{23} (\bibinfo{year}{2017}) \bibinfo{pages}{195--203}.
%Type = Article
\bibitem[{Osaba et~al.(2017)Osaba, Yang, Diaz, Onieva, Masegosa, and
  Perallos}]{Osaba2017FA}
\bibinfo{author}{E.~Osaba}, \bibinfo{author}{X.-S. Yang},
  \bibinfo{author}{F.~Diaz}, \bibinfo{author}{E.~Onieva},
  \bibinfo{author}{A.~Masegosa}, \bibinfo{author}{A.~Perallos},
\newblock \bibinfo{title}{A discrete firefly algorithm to solve a rich vehicle
  routing problem modelling a newspaper distribution system with recycling
  policy},
\newblock \bibinfo{journal}{Soft Computing} \bibinfo{volume}{21}
  (\bibinfo{year}{2017}) \bibinfo{pages}{5295--5308}.
%Type = Article
\bibitem[{Osaba et~al.(2019)Osaba, Yang, Jr., Lopez-Garcia, and
  Vazquez-Paravila}]{Osaba2019BA}
\bibinfo{author}{E.~Osaba}, \bibinfo{author}{X.-S. Yang},
  \bibinfo{author}{I.~F. Jr.}, \bibinfo{author}{P.~Lopez-Garcia},
  \bibinfo{author}{A.~Vazquez-Paravila},
\newblock \bibinfo{title}{A discrite and improved bat algorithm for solving a
  medical goods distribution problem with pharmacological waste collection},
\newblock \bibinfo{journal}{Swarm and Evolutionary Computation}
  \bibinfo{volume}{44} (\bibinfo{year}{2019}) \bibinfo{pages}{273--286}.
%Type = Article
\bibitem[{Rango et~al.(2018)Rango, Palmieri, Yang, and Marano}]{Rango2018}
\bibinfo{author}{F.~D. Rango}, \bibinfo{author}{N.~Palmieri},
  \bibinfo{author}{X.-S. Yang}, \bibinfo{author}{S.~Marano},
\newblock \bibinfo{title}{Swarm robotics in wireless distributed protocol
  design for coordinating robots invovled in cooperative tasks},
\newblock \bibinfo{journal}{Soft Computing} \bibinfo{volume}{22}
  (\bibinfo{year}{2018}) \bibinfo{pages}{4251--4266}.
%Type = Article
\bibitem[{Cagnina et~al.(2008)Cagnina, Esquivel, and Coello~Coello}]{Cag2008}
\bibinfo{author}{L.~C. Cagnina}, \bibinfo{author}{S.~C. Esquivel},
  \bibinfo{author}{A.~C. Coello~Coello},
\newblock \bibinfo{title}{Solving engineering optimization problems with the
  simple constrained particle swarm optimizer},
\newblock \bibinfo{journal}{Informatica} \bibinfo{volume}{32}
  (\bibinfo{year}{2008}) \bibinfo{pages}{319--326}.
%Type = Article
\bibitem[{Deb et~al.(2002)Deb, Pratap, Agarwal, and Mayarivan}]{Deb2002NSGA}
\bibinfo{author}{K.~Deb}, \bibinfo{author}{A.~Pratap},
  \bibinfo{author}{S.~Agarwal}, \bibinfo{author}{T.~Mayarivan},
\newblock \bibinfo{title}{A fast and elitist multiobjective algorithm:
  Nsga-ii},
\newblock \bibinfo{journal}{IEEE Transactions on Evolutionary Computation}
  \bibinfo{volume}{6} (\bibinfo{year}{2002}) \bibinfo{pages}{182--197}.
%Type = Article
\bibitem[{Yang and Gandomi(2012)}]{YangGandomi2012BA}
\bibinfo{author}{X.-S. Yang}, \bibinfo{author}{A.~H. Gandomi},
\newblock \bibinfo{title}{Bat algorithm: a novel approach for global
  engineering optimization},
\newblock \bibinfo{journal}{Engineering Computation} \bibinfo{volume}{29}
  (\bibinfo{year}{2012}) \bibinfo{pages}{464--483}.
%Type = Article
\bibitem[{Yang and Deb(2013)}]{Yang2013MOCS}
\bibinfo{author}{X.-S. Yang}, \bibinfo{author}{S.~Deb},
\newblock \bibinfo{title}{Multiobjective cuckoo search for design
  optimization},
\newblock \bibinfo{journal}{Computers \& Operations Research}
  \bibinfo{volume}{40} (\bibinfo{year}{2013}) \bibinfo{pages}{1616--1624}.
%Type = Article
\bibitem[{Gandom and Yang(2014)}]{Gandomi2014CBA}
\bibinfo{author}{A.~H. Gandom}, \bibinfo{author}{X.-S. Yang},
\newblock \bibinfo{title}{Chaotic bat algorithm},
\newblock \bibinfo{journal}{Journal of Computational Science}
  \bibinfo{volume}{5} (\bibinfo{year}{2014}) \bibinfo{pages}{224--232}.
%Type = Article
\bibitem[{Chakri et~al.(2017)Chakri, Khelif, Benouaret, and Yang}]{Chakri2017}
\bibinfo{author}{A.~Chakri}, \bibinfo{author}{R.~Khelif},
  \bibinfo{author}{M.~Benouaret}, \bibinfo{author}{X.-S. Yang},
\newblock \bibinfo{title}{New directional bat algorithm for continuous
  optimization problems},
\newblock \bibinfo{journal}{Expert Systems with Applications}
  \bibinfo{volume}{69} (\bibinfo{year}{2017}) \bibinfo{pages}{159--175}.
%Type = Article
\bibitem[{Papa et~al.(2017)Papa, Rosa, Pereira, and Yang}]{Papa2017Q}
\bibinfo{author}{J.~P. Papa}, \bibinfo{author}{G.~H. Rosa},
  \bibinfo{author}{D.~R. Pereira}, \bibinfo{author}{X.-S. Yang},
\newblock \bibinfo{title}{Quaternion-based deep belief networks fine-tuning},
\newblock \bibinfo{journal}{Applied Soft Computing} \bibinfo{volume}{60}
  (\bibinfo{year}{2017}) \bibinfo{pages}{328--335}.
%Type = Article
\bibitem[{Palmieri et~al.(2018)Palmieri, Yang, Rango, and
  Santamaria}]{Palmieri2018}
\bibinfo{author}{N.~Palmieri}, \bibinfo{author}{X.-S. Yang},
  \bibinfo{author}{F.~D. Rango}, \bibinfo{author}{A.~F. Santamaria},
\newblock \bibinfo{title}{Self-adaptive decision-making mechanisms to balance
  the execution of multiple tasks for a multi-robots team},
\newblock \bibinfo{journal}{Neurocomputing} \bibinfo{volume}{306}
  (\bibinfo{year}{2018}) \bibinfo{pages}{17--36}.
%Type = Article
\bibitem[{Ouaarab et~al.(2014)Ouaarab, Ahiod, and Yang}]{Ouaarab2014}
\bibinfo{author}{A.~Ouaarab}, \bibinfo{author}{B.~Ahiod},
  \bibinfo{author}{X.-S. Yang},
\newblock \bibinfo{title}{Discrite cuckoo search algorithm for the travelling
  salesman problem},
\newblock \bibinfo{journal}{Neural Computing and Applications}
  \bibinfo{volume}{24} (\bibinfo{year}{2014}) \bibinfo{pages}{1659--1669}.
%Type = Article
\bibitem[{Osaba et~al.(2016)Osaba, Yang, Diaz, Lopez-Garcia, and
  Carballedo}]{Osaba2016BA}
\bibinfo{author}{E.~Osaba}, \bibinfo{author}{X.-S. Yang},
  \bibinfo{author}{F.~Diaz}, \bibinfo{author}{P.~Lopez-Garcia},
  \bibinfo{author}{R.~Carballedo},
\newblock \bibinfo{title}{An improved discrete bat algorithm for symmetric and
  assymmetric travelling salesman problems},
\newblock \bibinfo{journal}{Engineering Applications of Artificial
  Intelligence} \bibinfo{volume}{48} (\bibinfo{year}{2016})
  \bibinfo{pages}{59--71}.
%Type = Article
\bibitem[{Blum and Roli(2003)}]{Blum2003}
\bibinfo{author}{C.~Blum}, \bibinfo{author}{A.~Roli},
\newblock \bibinfo{title}{Metaheuristics in combinatorial optimization:
  overview and conceptural comparison},
\newblock \bibinfo{journal}{ACM Comput. Survey} \bibinfo{volume}{25}
  (\bibinfo{year}{2003}) \bibinfo{pages}{268--308}.
%Type = Article
\bibitem[{Senthilnath et~al.(2011)Senthilnath, Omkar, and Mani}]{Senthil2011}
\bibinfo{author}{J.~Senthilnath}, \bibinfo{author}{S.~N. Omkar},
  \bibinfo{author}{V.~Mani},
\newblock \bibinfo{title}{Clustering using firefly algorithm: performance
  study},
\newblock \bibinfo{journal}{Swarm and Evolutionary Computation}
  \bibinfo{volume}{1} (\bibinfo{year}{2011}) \bibinfo{pages}{164--171}.
%Type = Book
\bibitem[{Yang and He(2019)}]{YangHe2019}
\bibinfo{author}{X.-S. Yang}, \bibinfo{author}{X.-S. He},
  \bibinfo{title}{Mathematical Foundations of Nature-Inspired Algorithms},
  Springer Briefs in Optimization, \bibinfo{publisher}{Springer},
  \bibinfo{address}{Cham, Switzerland}, \bibinfo{year}{2019}.
%Type = Book
\bibitem[{Grindstead and Snell(1997)}]{Grindstead1997}
\bibinfo{author}{C.~M. Grindstead}, \bibinfo{author}{J.~L. Snell},
  \bibinfo{title}{Introduction to Probability}, \bibinfo{publisher}{Americal
  Mathematical Society}, \bibinfo{address}{Providence, Rhode Island},
  \bibinfo{edition}{second} edition, \bibinfo{year}{1997}.
%Type = Book
\bibitem[{Bhat and Miller(2002)}]{Bhat2002}
\bibinfo{author}{U.~N. Bhat}, \bibinfo{author}{G.~K. Miller},
  \bibinfo{title}{Elements of Applied Stochastic Processes},
  \bibinfo{publisher}{John Wiley \& Sons}, \bibinfo{address}{New York},
  \bibinfo{edition}{third} edition, \bibinfo{year}{2002}.
%Type = Book
\bibitem[{Gutowski(2001)}]{Gutow2001}
\bibinfo{author}{M.~Gutowski}, \bibinfo{title}{L\'evy flights as an undelying
  mechanism for global optimization algorithms}, \bibinfo{publisher}{ArXiv
  Mathematical Physics e-Prints}, \bibinfo{address}{Accessed 1 Sept 2019},
  \bibinfo{year}{2001}.
%Type = Article
\bibitem[{Clerc and Kennedy(2002)}]{Clerc2002}
\bibinfo{author}{M.~Clerc}, \bibinfo{author}{J.~Kennedy},
\newblock \bibinfo{title}{The particle swarm: explosion, stability, and
  convergence in a multidimensional complex space},
\newblock \bibinfo{journal}{IEEE Transactions on Evolutionary Computation}
  \bibinfo{volume}{6} (\bibinfo{year}{2002}) \bibinfo{pages}{58--73}.
%Type = Article
\bibitem[{Chen et~al.(2018)Chen, Peng, Xing-Shi, and Yang}]{Chen2018}
\bibinfo{author}{S.~Chen}, \bibinfo{author}{G.-H. Peng},
  \bibinfo{author}{Xing-Shi}, \bibinfo{author}{X.-S. Yang},
\newblock \bibinfo{title}{Global convergence analysis of the bat algorithm
  using a markovian framework and dynamic system theory},
\newblock \bibinfo{journal}{Expert Systems with Applications}
  \bibinfo{volume}{114} (\bibinfo{year}{2018}) \bibinfo{pages}{173--182}.
%Type = Book
\bibitem[{Granas and Dugundji(2003)}]{Granas2003}
\bibinfo{author}{A.~Granas}, \bibinfo{author}{J.~Dugundji},
  \bibinfo{title}{Fixed Point Theory}, \bibinfo{publisher}{Springer-Verlag},
  \bibinfo{address}{New York}, \bibinfo{year}{2003}.
%Type = Book
\bibitem[{Khamsi and Kirk(2001)}]{Khamsi2001Kirk}
\bibinfo{author}{M.~A. Khamsi}, \bibinfo{author}{W.~A. Kirk},
  \bibinfo{title}{An Introduction to Metric Space and Fixed Point Theory},
  \bibinfo{publisher}{John Wiley \& Sons}, \bibinfo{address}{New York},
  \bibinfo{year}{2001}.
%Type = Book
\bibitem[{Khalil(1996)}]{Khalil1996}
\bibinfo{author}{H.~Khalil}, \bibinfo{title}{Nonlinear Systems},
  \bibinfo{publisher}{Prentice Hall}, \bibinfo{address}{New Jersey},
  \bibinfo{edition}{third} edition, \bibinfo{year}{1996}.
%Type = Article
\bibitem[{Suzuki(1995)}]{Suzuki1995}
\bibinfo{author}{J.~Suzuki},
\newblock \bibinfo{title}{A markov chain analysis on simple genetic
  algorithms},
\newblock \bibinfo{journal}{IEEE Trans Sys Man Cybern} \bibinfo{volume}{25}
  (\bibinfo{year}{1995}) \bibinfo{pages}{655--659}.
%Type = Article
\bibitem[{Aytug et~al.(1996)Aytug, Bhattacharrya, and Koehler}]{Aytug1996}
\bibinfo{author}{H.~Aytug}, \bibinfo{author}{S.~Bhattacharrya},
  \bibinfo{author}{G.~J. Koehler},
\newblock \bibinfo{title}{A markov chain analysis of genetic algorithms with
  power of 2 cardinality alphabets},
\newblock \bibinfo{journal}{European Journal of Operational Research}
  \bibinfo{volume}{96} (\bibinfo{year}{1996}) \bibinfo{pages}{195--201}.
%Type = Article
\bibitem[{Greenhalgh and Marshal(2000)}]{Greenh2000}
\bibinfo{author}{D.~Greenhalgh}, \bibinfo{author}{S.~Marshal},
\newblock \bibinfo{title}{Convergence criteria for genetic algorithm},
\newblock \bibinfo{journal}{SIAM Journal Comput} \bibinfo{volume}{30}
  (\bibinfo{year}{2000}) \bibinfo{pages}{269--282}.
%Type = Article
\bibitem[{Gutjahr(2010)}]{Gutj2010}
\bibinfo{author}{W.~J. Gutjahr},
\newblock \bibinfo{title}{Convergence analysis of metaheurtics},
\newblock \bibinfo{journal}{Annalysis Inf Sys} \bibinfo{volume}{10}
  (\bibinfo{year}{2010}) \bibinfo{pages}{159--187}.
%Type = Incollection
\bibitem[{He et~al.(2018)He, Wang, Wang, and Yang}]{He2018CS}
\bibinfo{author}{X.~S. He}, \bibinfo{author}{F.~Wang},
  \bibinfo{author}{Y.~Wang}, \bibinfo{author}{X.~S. Yang},
\newblock \bibinfo{title}{Global convergence analysis of cuckoo search using
  markov theory},
\newblock in: \bibinfo{editor}{X.-S. Yang} (Ed.),
  \bibinfo{booktitle}{Nature-Inspired Algorithms and Applied Optimization},
  volume \bibinfo{volume}{744}, \bibinfo{publisher}{Springer Nature},
  \bibinfo{address}{Cham, Switzerland}, \bibinfo{year}{2018}, pp.
  \bibinfo{pages}{53--67}.
%Type = Article
\bibitem[{Ghate and Smith(2008)}]{Ghate2008}
\bibinfo{author}{A.~Ghate}, \bibinfo{author}{R.~Smith},
\newblock \bibinfo{title}{Adaptive search with stochastic acceptance
  probability for global optimization},
\newblock \bibinfo{journal}{Operations Research Letters} \bibinfo{volume}{36}
  (\bibinfo{year}{2008}) \bibinfo{pages}{285--290}.
%Type = Book
\bibitem[{Bertsekas et~al.(2003)Bertsekas, Nedic, and Ozdaglar}]{Berts2003}
\bibinfo{author}{D.~P. Bertsekas}, \bibinfo{author}{A.~Nedic},
  \bibinfo{author}{A.~Ozdaglar}, \bibinfo{title}{Convex Analysis and
  Optimization}, \bibinfo{publisher}{Athena Scientific},
  \bibinfo{address}{Belmont, MA}, \bibinfo{edition}{second} edition,
  \bibinfo{year}{2003}.
%Type = Book
\bibitem[{Chabert(1999)}]{Chabert1999}
\bibinfo{author}{J.~L. Chabert}, \bibinfo{title}{A History of Algorithms: From
  the Pebble to the Microchips}, \bibinfo{publisher}{Springer-Verlag},
  \bibinfo{address}{Heidelberg}, \bibinfo{year}{1999}.
%Type = Book
\bibitem[{Zdenek(2009)}]{Zdenek2009}
\bibinfo{author}{D.~Zdenek}, \bibinfo{title}{Optimal Quadratic Programming
  Algorithms: With Applications to Variational Inequalities},
  \bibinfo{publisher}{Springer}, \bibinfo{address}{Heidelberg},
  \bibinfo{year}{2009}.
%Type = Article
\bibitem[{Eiben and Smit(2011)}]{Eiben2011}
\bibinfo{author}{A.~E. Eiben}, \bibinfo{author}{S.~K. Smit},
\newblock \bibinfo{title}{Parameter tuning for configuring and analyzing
  evolutionary algorithms},
\newblock \bibinfo{journal}{Swarm and Evolutionary Computation}
  \bibinfo{volume}{1} (\bibinfo{year}{2011}) \bibinfo{pages}{19--31}.
%Type = Article
\bibitem[{Brest et~al.(2006)Brest, Greiner, Boskovic, Mernik, and
  Zumer}]{Brest2006DE}
\bibinfo{author}{J.~Brest}, \bibinfo{author}{S.~Greiner},
  \bibinfo{author}{B.~Boskovic}, \bibinfo{author}{M.~Mernik},
  \bibinfo{author}{V.~Zumer},
\newblock \bibinfo{title}{Self-adapting control parameters in differential
  evolution: a comparative study on numerical benchmark functions},
\newblock \bibinfo{journal}{IEEE Transactions on Evolutionary Computation}
  \bibinfo{volume}{10} (\bibinfo{year}{2006}) \bibinfo{pages}{646--657}.
%Type = Article
\bibitem[{Yang et~al.(2013)Yang, Deb, Loomes, and Karamanoglu}]{Yang2013STA}
\bibinfo{author}{X.-S. Yang}, \bibinfo{author}{S.~Deb},
  \bibinfo{author}{M.~Loomes}, \bibinfo{author}{M.~Karamanoglu},
\newblock \bibinfo{title}{A framework for self-tuning optimization algorithm},
\newblock \bibinfo{journal}{Neural Computing and Applications}
  \bibinfo{volume}{23} (\bibinfo{year}{2013}) \bibinfo{pages}{2051--2057}.
%Type = Article
\bibitem[{Jamil and Yang(2013)}]{Jamil2013}
\bibinfo{author}{M.~Jamil}, \bibinfo{author}{X.-S. Yang},
\newblock \bibinfo{title}{A literature survey of benchmark functions for global
  optimisation problems},
\newblock \bibinfo{journal}{International Journal of Mathematical Modelling and
  Numerical Optimisation} \bibinfo{volume}{4} (\bibinfo{year}{2013})
  \bibinfo{pages}{150--194}.
%Type = Article
\bibitem[{Wolpert and Macready(1997)}]{Wolpert1997}
\bibinfo{author}{D.~H. Wolpert}, \bibinfo{author}{W.~G. Macready},
\newblock \bibinfo{title}{No free lunch theorems for optimization},
\newblock \bibinfo{journal}{IEEE Treansactions on Evolutionary Computation}
  \bibinfo{volume}{1} (\bibinfo{year}{1997}) \bibinfo{pages}{67--82}.
%Type = Incollection
\bibitem[{Joyce and Herrmann(2018)}]{Joyce2018}
\bibinfo{author}{T.~Joyce}, \bibinfo{author}{J.~M. Herrmann},
\newblock \bibinfo{title}{A review of no free lunch theorems, and their
  implicatoins for metaheuristic optimisation},
\newblock in: \bibinfo{editor}{X.-S. Yang} (Ed.),
  \bibinfo{booktitle}{Nature-Inspired Algorithms and Applied Optimization},
  \bibinfo{publisher}{Springer}, \bibinfo{address}{Cham, Switzerland},
  \bibinfo{year}{2018}, pp. \bibinfo{pages}{27--52}.
%Type = Article
\bibitem[{Auger and Teytaud(2010)}]{Auger2010FL}
\bibinfo{author}{A.~Auger}, \bibinfo{author}{O.~Teytaud},
\newblock \bibinfo{title}{Continuous lunches are free plus the design of
  optimal optimization algorithms},
\newblock \bibinfo{journal}{Algorithmica} \bibinfo{volume}{57}
  (\bibinfo{year}{2010}) \bibinfo{pages}{121--146}.
%Type = Article
\bibitem[{Corne and Knowles(2003)}]{Corne2003FL}
\bibinfo{author}{D.~Corne}, \bibinfo{author}{J.~Knowles},
\newblock \bibinfo{title}{Some multiobjective optimizers are better than
  others},
\newblock \bibinfo{journal}{Evolutionary Computation} \bibinfo{volume}{4}
  (\bibinfo{year}{2003}) \bibinfo{pages}{2506--2512}.
%Type = Article
\bibitem[{Wolpert and Macready(2005)}]{Wolpert2005}
\bibinfo{author}{D.~H. Wolpert}, \bibinfo{author}{W.~G. Macready},
\newblock \bibinfo{title}{Coevolutionary free lunches},
\newblock \bibinfo{journal}{IEEE Transactions on Evolutionary Computation}
  \bibinfo{volume}{9} (\bibinfo{year}{2005}) \bibinfo{pages}{721--735}.
%Type = Article
\bibitem[{Cook(1983)}]{Cook1983}
\bibinfo{author}{S.~Cook},
\newblock \bibinfo{title}{An overview of computational complexity},
\newblock \bibinfo{journal}{Commun. ACM} \bibinfo{volume}{26}
  (\bibinfo{year}{1983}) \bibinfo{pages}{400--408}.
%Type = Book
\bibitem[{Arara and Barak(2009)}]{Arara2009}
\bibinfo{author}{S.~Arara}, \bibinfo{author}{B.~Barak},
  \bibinfo{title}{Computational Complexity: A Modern Approach},
  \bibinfo{publisher}{Cambridge University Press}, \bibinfo{address}{Cambridge,
  UK}, \bibinfo{year}{2009}.
%Type = Book
\bibitem[{Goldreich(2008)}]{Goldreich2008}
\bibinfo{author}{O.~Goldreich}, \bibinfo{title}{Computational Complexity: A
  Conceptual Perspective}, \bibinfo{publisher}{Cambridge University Press},
  \bibinfo{address}{Cambridge, UK}, \bibinfo{year}{2008}.

\end{thebibliography}
\end{document}